\documentclass[letterpaper, 10 pt, conference]{ieeeconf}  % Comment this line out if you need a4paper

\IEEEoverridecommandlockouts                              % This command is only needed if 
\usepackage{color,soul}
\usepackage{hyperref}
\usepackage{mwe}
\usepackage{graphics} % for pdf, bitmapped graphics files
\usepackage{epsfig} % for postscript graphics files
\usepackage{mathptmx} % assumes new font selection scheme installed
\usepackage{times} % assumes new font selection scheme installed
\usepackage{amsmath} % assumes amsmath package installed
\usepackage{amssymb}  % assumes amsmath package installed
\usepackage{subfigure}
\usepackage[noadjust]{cite}
\usepackage{booktabs}
\usepackage{multirow}
\usepackage{algorithmic}
\usepackage[ruled,norelsize]{algorithm2e} % DO NOT USE FLOATING
\def\ie{\emph{i.e.}} 
\def\eg{\emph{e.g.}} 

\title{\LARGE \bf
Mobile MoCap: Retroreflector Localization On-The-Go
}

\author{Gary Lvov$^{*}$, Mark Zolotas, Nathaniel Hanson, Austin Allison,\\ Xavier Hubbard, Michael Carvajal, Ta\c{s}kin Padir% <-this % stops a space
\thanks{Institute for Experiential Robotics, Northeastern University, Boston, MA, USA}%}
\thanks{*Corresponding author: {\tt\small lvov.g@northeastern.edu}}%}
}

\begin{document}

\maketitle
\thispagestyle{empty}
\pagestyle{empty}

%%%%%%%%%%%%%%%%%%%%%%%%%%%%%%%%%%%%%%%%%%%%%%%%%%%%%%%%%%%%%%%%%%%%%%%%%%%%%%%%
\begin{abstract}
Motion capture through tracking retroreflectors obtains highly accurate pose estimation, which is frequently used in robotics. Unlike commercial motion capture systems, fiducial marker-based tracking methods, such as AprilTags, can perform relative localization without requiring a static camera setup. However, popular pose estimation methods based on fiducial markers have lower localization accuracy than commercial motion capture systems. We propose Mobile MoCap, a system that utilizes inexpensive near-infrared cameras for accurate relative localization even while in motion. We present a retroreflector feature detector that performs 6-DoF (six degrees-of-freedom) tracking and operates with minimal camera exposure times to reduce motion blur. To evaluate the proposed localization technique while in motion, we mount our Mobile MoCap system, as well as an RGB camera to benchmark against fiducial markers, onto a precision-controlled linear rail and servo. The fiducial marker approach employs AprilTags, which are pervasively used for localization in robotics. We evaluate the two systems at varying distances, marker viewing angles, and relative velocities. Across all experimental conditions, our stereo-based Mobile MoCap system obtains higher position and orientation accuracy than the fiducial approach.

The code for Mobile MoCap is implemented in ROS 2 and made publicly available at~\url{https://github.com/RIVeR-Lab/mobile_mocap}.

\end{abstract}

%%%%%%%%%%%%%%%%%%%%%%%%%%%%%%%%%%%%%%%%%%%%%%%%%%%%%%%%%%%%%%%%%%%%%%%%%%%%%%%%
\section{Introduction}

Localization of people and robots is a critical need for the future of automation. Recognizing the pose of dynamic objects enables a robot to not only understand its current location in an environment, but also to ensure an appropriate factor of safety. In settings where robots and humans occupy a shared space, safety is paramount and the margin for error has to be minimized. Artificial landmarks, such as fiducial markers, provide an easily identifiable reference point for cameras to track objects in 3D space~\cite{Olson2011April,Fiala2005ARtag,Wang2016April2}. Recognizable marker geometries are used to localize fiducial markers and to discriminate between objects. This technique has applications in virtual reality~\cite{simonetto2022methodological}, medicine~\cite{alarcon2020upper}, and robotics~\cite{chen2021autonomous,Zolotas2018,pfrommer2019tagslam,Kalaitzakis2021Fiducial}.

\begin{figure}[t]
    \centering
    \includegraphics[scale=.25]{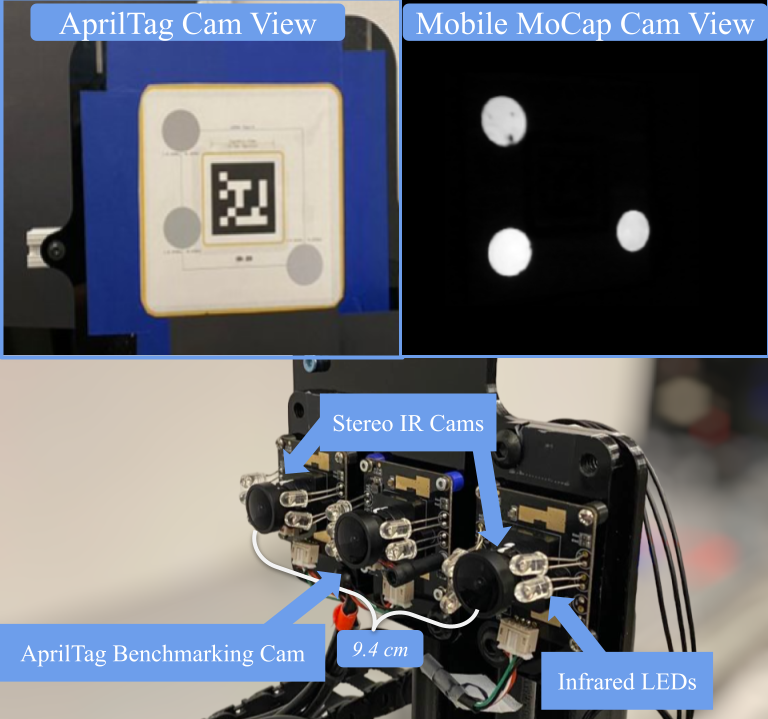}
    \caption{Testing setup to evaluate a fiducial marker approach, AprilTags, against the proposed Mobile MoCap. The central camera is in RGB mode for AprilTag tracking. The outer two cameras are in IR mode for stereo Mobile MoCap. All three cameras are the same model for a fair evaluation.}
    \label{fig:title_teaser}
    \vspace{-2.2mm}
\end{figure}

In robotics, relative localization of artificial landmarks or fiducials is used to perform visual servoing~\cite{Kalaitzakis2021Fiducial}. This has additional applications to autonomous vehicles, where landmarks and fiducials can be used to provide a static reference point for simultaneous localization and mapping~\cite{Pfrommer2019TagSLAMRS}. However, common conditions that arise in robotics applications, such as motion blur, changes in perceived lighting, large viewing angles, and large distances, can all decrease the accuracy of contemporary approaches. Poor relative localization leads to erroneous robot movement when vision informs control, resulting in potentially dangerous situations.

To address the limitations of current fiducial-based approaches, we present Mobile MoCap, a stereo-based localization approach that uses retroreflectors, as inspired by commercial motion capture systems~\cite{Field2009Motion}. Fig.~\ref{fig:title_teaser} demonstrates an overview of our proposed system. Retroreflectors are commercially available as either flat, cut-to-size, textured sheets of adhesive-backed paper, or as screw-mounted, textured spheres of plastic. Affixing three of these retroreflectors to the same rigid target body is all that is needed to extract said target's 6D (six degrees-of-freedom,~\ie, 6-DoF) pose by triangulating the unique 3D positions of each of the attached retroreflectors. These retroreflective markers reflect near infrared (NIR) light into an NIR camera, making them easily identifiable under dim lighting conditions, or even within scenes of complete darkness. Furthermore, unlike 2D fiducial markers, retroreflectors can be arranged in a non-planar layout, making retroreflector-based detection systems more robust to a wider range of IR camera viewing angles.

Mobile MoCap has a twofold benefit over fiducial markers. First, by decreasing the exposure time of the cameras, we inherently reduce motion blur while still being able to isolate retroreflectors from the scene's background. Second, by not being constrained to planar mounting surfaces when distributing retroreflectors over the target object's unique geometry, we can achieve better target tracking in 3D space.

We designed a robotic experimental testbed to autonomously benchmark our Mobile MoCap system against AprilTags~\cite{Olson2011April,Wang2016April2}, a popular fiducial marker tracking approach. The robotic testbed consists of a camera assembly mounted onto a servo and actuated linear rail, enabling each camera's viewing angle and distance to be controlled relative to a static target. In our experiments, we consider different viewing angles, distances, and relative velocities to a target. The target contains both retroreflectors and an AprilTag for the purpose of comparing the marker localization accuracy of the two tracking techniques. The overall setup was designed to mimic a scenario in which a mobile robot aims to localize a static object of interest, such as a charging station or a worker wearing a reflective safety vest.

The key contributions of this paper include:
\begin{itemize}
    \item A lightweight, inexpensive visual localization stereo system capable of real-time 6-DoF tracking, which succeeds under dynamic conditions that are challenging to contemporary fiducial marker-based approaches, while offering superior accuracy.
    \item A robotics testing platform to benchmark our proposed Mobile MoCap system against a pervasive fiducial marker baseline, where the cameras are mounted onto an actuated rail and perform relative localization at various distances, viewing angles, and relative velocities.
    \item An open-source software framework implemented atop the Robot Operating System (ROS) 2~\cite{Macenski2022ROS2} and OpenCV~\cite{opencv_library} for easy interoperability with robotic systems and inexpensive camera hardware.
\end{itemize}

\section{Related Work}

There are numerous approaches to vision-based localization and tracking of objects within an environment. One prominent approach is to track an object directly, without using markers or any other landmarks. These ``markerless'' approaches are attractive because they do not depend on extra hardware for tracking~\cite{li2022bcot}. For example, Region-based Convolutional Neural Networks (R-CNNs) can perform markerless tracking by combining regional classification with CNNs~\cite{Girshick2014RCNN}. Nevertheless, learning-based markerless pose estimation requires extensive training data and does not yet demonstrate localization accuracy comparable to those of marker-based approaches~\cite{Nakano2020Evaluation}. 

In contrast, marker-based tracking methods rely on visually distinct landmarks that can be easily identified by an optical camera. Active markers are a common example, which require a power supply and supporting electronics to function,~\eg, pulsed RGB LEDs during human hand tracking~\cite{Ruppel2019Low}. These markers excel under low-lighting conditions, but the requirement for a power source makes them more resource-intensive than other solutions. Another example are low-powered retroreflector tags that can be localized at larger distances~\cite{Soltanaghaei2021Millimetro}. These tags are favored in high-speed vehicle applications, despite the need for supporting hardware.

Passive markers are an alternative solution to remedy the issue of requiring additional electronics or a power supply, thus offering a lightweight cost-effective tracking approach. For instance, passive markers are a preferable choice in full-body motion capture as they ensure that the participant is not encumbered by supporting hardware~\cite{Field2009Motion}. One drawback of passive markers is that they need direct line of sight with a camera, hence occlusions can prevent effective tracking. Commercial motion capture rigs solve this issue by leveraging multiple cameras to continuously track a single marker from all viewpoints. However, the static configuration of cameras around the indoor environment for commercial motion capture systems does not always translate well to robotics, as cameras in robotics are often mounted at fixed positions on the robot's body and facing outwards.

Another common solution to marker-based tracking that is widely used in robotics are fiducial markers~\cite{Kalaitzakis2021Fiducial}. Fiducials are visually distinct, flat, black and white printed images,~\eg, AprilTag~\cite{Olson2011April} or ARTag~\cite{Fiala2005ARtag}. These fiducials are easy and quick to localize, even when using low resolution images that are partially occluded or are not orthogonal to the camera’s line of sight~\cite{Olson2011April,Wang2016April2}. Fiducial markers have a plethora of applications in robotics, most notably in visual servoing and autonomous vehicles~\cite{Pfrommer2019TagSLAMRS}. 

Passive fiducial tracking methods can be improved by incorporating retroreflective material and multiple cameras into the setup. Monocular camera setups are lower cost and less space obtrusive~\cite{Vagvolgyi2022Wide,Lichtenauer2011Monocular,Smith2021Method}. However, stereo cameras offer distinct advantages, such as higher localization accuracy and more effective acquisition of scene information~\cite{Azad2009Stereo}. Stereo motion capture methods provide high precision localization useful for robotics~\cite{MOLLER2016389}. In prior work, stereo cameras with a wide field of view (FoV) and an occlusion detection algorithm were shown to rival commercial motion capture systems, albeit not in real-time~\cite{Islam2020Stereo}. Trinocular setups with NIR cameras have also demonstrated the capacity to track passive markers at sub-tenth millimeter precision~\cite{Bi2021High}. Nonetheless, a multiple camera system can rapidly become expensive and typically remains stationary. 

Few existing marker-based tracking systems are mobile, while retaining the localization accuracy of commercial motion capture technology. Previous work in augmented reality for surgical use has utilized the stereo cameras on the Microsoft HoloLens to track a surgical tool of known length~\cite{Gsaxner2021Inside}. The study's results were reported at sub-centimeter precision under this mobile stereo configuration. We introduce a more general system that can have widespread use in robotics, where stereo cameras must handle motion at significantly larger speeds and distances than in augmented reality use-cases.

\section{Stereo-based Mobile MoCap}

Given the prevalence of the ROS 2~\cite{Macenski2022ROS2} middleware, we implemented the entire software and hardware stack of Mobile MoCap with the goals of being: 1) open-source; 2) cross-platform; and 3) easily reproducible.

\subsection{System Hardware}

A stereo NIR camera setup with an integrated light source can be constructed from inexpensive off-the-shelf components. ArduCam\footnote{\url{https://www.arducam.com/product-category/uvc-usb-camera-module/usb-uvc-cameras-night-vision/}} offers an NIR USB camera that provides a camera stream running at 30 frames per second (FPS) and 640x480p resolution. When properly focused, the cameras have an FoV of $100^{\circ}$ horizontal and $138^{\circ}$ diagonal. These cameras utilize a silicon-based photodetector with broadband sensitivity. We permanently enable the IR cut filter by removing the transition photoresistor to allow for IR tracking in well-illuminated indoor environments, which would normally cause the camera to autoswitch to the RGB space. IR LEDs adjacent to the camera lens provide illumination at 850~nm, within the rated tolerance of the photodetector, but outside human visual range. Creating a stereo NIR camera from the ArduCams costs less than \$100, including \$20 in materials to fix the two cameras together. For later benchmarking against AprilTag, a third unmodified ArduCam of the same model is operated in the RGB space. 
% A full parts manifest and assembly instructions are available on the project website.

The monocular and stereo camera subsystems are combined in series on a laser cut acrylic plate to enable benchmarking of AprilTags against the proposed system. Tags are mounted on a continuous rotation servo (Dynamixel AX-12A). The entire construction, including the AprilTag benchmarking camera, weighs less than 250~grams and is less than 20~cm wide.

\subsection{ROS 2 Software Package}

\begin{figure}[t]
    \centering
    \vspace{4 mm} 
    \includegraphics[width=\columnwidth]{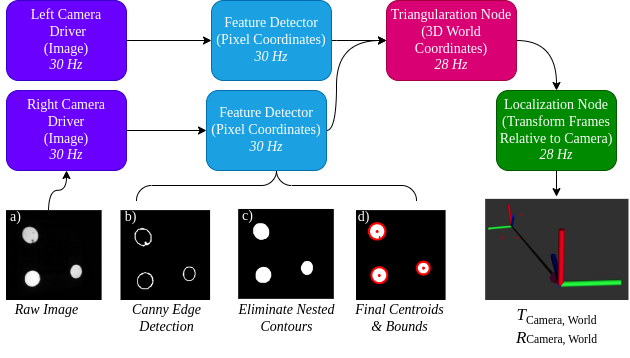}
    \caption{Feature detection pipeline deconstructed as a series of ROS 2 nodes. \textbf{(a)} A raw camera image of four retroreflectors with exposure time minimized; \textbf{(b)} Detection of contrasting edges on the retroreflectors; \textbf{(c)} Elimination of nested contours within the detected region; and \textbf{(d)} the final detected contours with their centroids.}
    \label{fig:ros2}
    \vspace{-1.9mm}
\end{figure}

We have developed a ROS 2 package~\cite{Macenski2022ROS2} to facilitate 6-DoF tracking (shown in Fig.~\ref{fig:ros2}) of rigid retroreflector geometry, as well as 3D translation estimation of individual retroreflectors. ROS 2 ensures easy interoperability with other robotics systems and has a rich set of transform libraries that are also suitable for geometric data processing. As detailed in Fig.~\ref{fig:ros2}, the core algorithm steps are encapsulated as ROS 2 nodes, each occupying an independent thread capable of running in parallel. Any image processing steps that exploit OpenCV library~\cite{opencv_library} functions are developed in C++ for performance gains, while the pose extraction and utility scripts are implemented in Python. To the authors' best knowledge, there is no open-source, publicly available stereo retroreflector tracking package for ROS 2. 

\subsection{Filtering and Feature Detection}

The two stereo cameras are initialized with their exposure times minimized (500~$\mu s$). The resulting images are predominantly occupied by retroreflectors and small sections of reflected light noise, as visualized in Fig.~\ref{fig:ros2}a. When the 750~nm IR cut filter is engaged, light sources other than the co-located IR LEDs are easily filtered out, as they appear distinctly green in IR space. Finally, a global intensity threshold is applied to the image, which can be tuned for the desired operating conditions. 

For every incoming frame, the pixel coordinates of each potential marker center are determined after filtering. The feature detector determines edges in the image using a Canny Edge Detector (Fig.~\ref{fig:ros2}b)~\cite{canny1986computational}. Multiple nested detections are corrected for by infilling the selected contours (Fig.~\ref{fig:ros2}c), resulting in the removal of redundant marker locations. Filling in nested contours helps prevent a single marker from being detected as more than one marker. Following the infilling process, an ellipse is fit to each contour. The marker radius is approximated as the average of the major and minor axes of the ellipse, with the ellipse's center as the feature location. Fitting an ellipse to the processed contours provides a better estimation of marker centers when there are partial occlusions than other techniques, such as fitting a minimum enclosing circle or applying the circle Hough Transform~\cite{hough1962method}. Yet in situations where markers are completely occluded, maintaining consistent tracking is difficult. This issue will be tackled in future work by leveraging a greater number of markers per tracked object to ensure reliable identification despite total occlusion of some markers. The final feature detection results are shown in Fig.~\ref{fig:ros2}d.

\subsection{Correspondence Matching}
\begin{algorithm}[t]
    \label{alg:corr}
    \caption{Geometric Correspondence Matching}
    \textbf{Inputs}: pixel coordinates $(x_i, y_i)  \; \forall$ features $i \in$ left camera; pixel coordinates $(x_j, y_j) \; \forall$ features $j \in$ right camera; \\
    \textbf{Initialize}: correspondences $\leftarrow$ []; assignments $\leftarrow$ [];
    
    \begin{algorithmic}
    \STATE{// find offset between one and every other feature $\forall i, j$ \\  }
    \STATE{geoms$_i$ $\leftarrow$ $\forall$ features $i \, (x_i, y_i)$.$offset(\forall$ other features)}
    \STATE{geoms$_j$ $\leftarrow$ $\forall$ features $j \, (x_j, y_j)$.$offset(\forall$ other features)}
    \STATE{$sort$(geoms$_i$); $sort$(geoms$_j$)}
    \STATE{shortest $\leftarrow$ $min$($length$($ $geoms$_i$), $length$(geoms$_j$))}
    \FOR{geometry$_i$ $\in$ geoms$_i$}
        \FOR {geometry$_j$  $\in$ geoms$_j$}
            \STATE{error $\leftarrow$ $||$geometry$_i$ - geometry$_j$}$||$
            \STATE{assignments.$append$(($i$, $j$), error)}
        \ENDFOR
    \ENDFOR
    \STATE{ $sort$(assignments) by increasing error}
    %\STATE{set$_i$ $\leftarrow \emptyset$, set$_j$ $\leftarrow \emptyset$}
    
    \FOR{assignment in assignments}
        \IF{features $i, j\in$ assignment not encountered} 
            \STATE{correspondences.$append$(($i, j$))}
        \ENDIF

        \IF{$length$(correspondences) $=$ shortest} 
            \RETURN {correspondences}
        \ENDIF
    \ENDFOR
    \RETURN{ambiguous correspondence}
    \end{algorithmic}
    % \begin{algorithmic}
    % \STATE{Input: pixel coordinates $(x_i, y_i)  \forall$ features $i \in$ left camera}
    % \STATE{pixel coordinates $(x_j, y_j) \forall$ features $j \in$ right camera}
    % \STATE{correspondence $\leftarrow$ []}
    % \STATE{assignments $\leftarrow$ []}
    % \STATE{geoms$_i$ $\leftarrow$ $\forall$ features $\in i$ $(x, y)$ offset to $\forall$ features $\in i$}
    % \STATE{geoms$_j$ $\leftarrow$ $\forall$ features $\in j$ $(x, y)$ offset to $\forall$ features $\in j$}
    % \STATE{$sort$(geoms$_i$), $sort$(geoms$_j$)}
    % \STATE{shortest $\leftarrow$ $min$($ length$($ $geoms$_i$), $length$(geoms$_j$))}
    % \FOR{geometry$_i$ $\in$ geoms$_i$}
    %     \FOR {geometry$_j$  $\in$ geoms$_j$}
    %         \STATE{error $\leftarrow$ $||$geometry$_i$ - geometry$_j$}$||$
    %         \STATE{assignments.$append$(($i$, $j$), error)}
    %     \ENDFOR
    % \ENDFOR
    % \STATE{ $sort$(assignments) by increasing error}
    % %\STATE{set$_i$ $\leftarrow \emptyset$, set$_j$ $\leftarrow \emptyset$}
    
    % \FOR{assignment in assignments}
    %     \IF{features $i, j\in$ assignment not yet encountered} 
    %         \STATE{correspondence.$append$(($i, j$))}
    %     \ENDIF

    %     \IF{$length$(correspondence) $=$ shortest} 
    %         \RETURN {correspondence}
    %     \ENDIF
    % \ENDFOR
    % \RETURN{ambiguous correspondence} 
    % \end{algorithmic}
    \label{multi_point}
\end{algorithm}

Once the marker centers are determined for both frames, the 3D locations of features are triangulated from corresponded feature pairs and the camera intrinsics/extrinsics. A core benefit of stereo-based tracking is that the 3D location of a feature can be triangulated from a single 2D feature pair, which is not necessarily correlated to other features in the scene. Camera projection matrices are determined by a stereo calibration.

% \begin{figure}[t]
%     \centering
%     \includegraphics[width=\columnwidth]{figures/assoc_cropped.png}
%     \caption{geometric alignment of features based on minimizing pixel-to-pixel distances.}
%     \label{fig:feature_corres}
% \end{figure}
An alignment method is used to match feature correspondences based on relative geometries, as summarized in Algorithm~\ref{alg:corr}. For each feature in the first frame, the offset to every other feature is calculated and stored as a relative geometry. For each feature in the second frame, the association that minimizes the pixel difference for that feature's relative geometry to the rest of the features is then found. Only one correspondence is assumed to be correct, so if there are several correspondences from a feature in the first frame to features in the second frame, only the association with the lowest pixel error is considered. This geometric matching approach is predicated on the assumption that the cameras have roughly the same orientation, similar extrinsics, and are not far enough apart to have vastly different views of features. Geometric matching is used due to the sparsity of the image. Using RANSAC~\cite{fischler1981random} to estimate the homography between the two images fails to determine correspondences, as keypoint detectors, such as ORB~\cite{rublee2011orb} and SURF~\cite{bay2008speeded}, do not locate a sufficient number of keypoints to correctly determine correspondence in sparse low exposure time images.

Once the feature correspondence is found, it is used along with the cameras' projection matrices and detected features to iteratively triangulate the 3D locations of the features. This approach is akin to the Iterative-LS method described by~\cite{Hartley1997Triangulation}. Triangulation allows a marker's 3D translation to be tracked by the cameras individually. 

Extracting multiple object poses from 3D marker locations is left for future work, as our current approach can only track a single rigid geometry comprised of three markers arranged in a scalene triangle. This is due to the fact that when there are multiple rigid bodies, it is non-trivial to associate the 3D points that belong to a specific rigid body, especially when the 3D features belonging to rigid bodies are occluded. 

The points are corresponded to a known stored geometry through an exhaustive search that correlates the longest edge between points to the longest edge in the stored scalene triangle. Once the point correspondence is determined, the translation and rotation are determined with a least-squares correspondence of two 3D point sets~\cite{arun1987least}.

\section{Experiments}

\begin{figure}[t]
    \centering
    \includegraphics[width=\columnwidth]{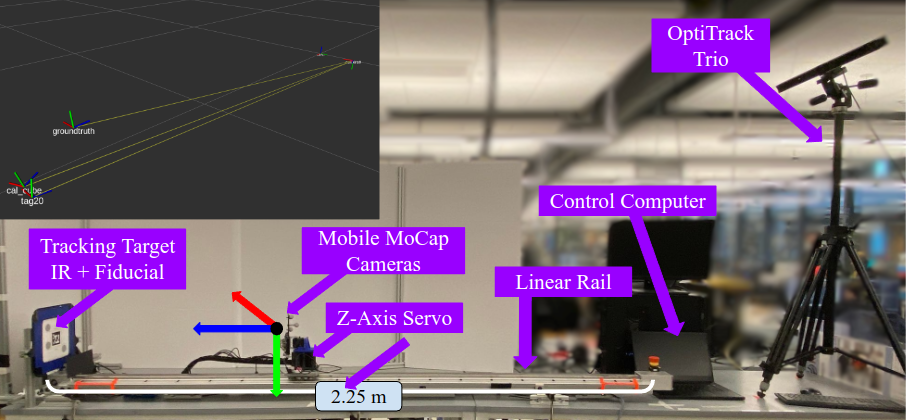}
    \caption{Experimental setup to benchmark the performance of the linear rail system against visual fiducial markers. Mobile MoCap cameras localize themselves relative to the tracking target, as does an AprilTag camera. Groundtruth is provided by the OptiTrack Trio, which determines the relative pose of the target and cameras. \textit{Inset}: Relative pose transforms from the motion capture setup.}
    \label{fig:experimental_setup}
    \vspace{-3.5mm}
\end{figure}

To evaluate our Mobile MoCap system, we designed a series of experiments using AprilTags with a monocular camera as a benchmark. An overview of the experimental setup is illustrated in Fig.~\ref{fig:experimental_setup}.

\subsection{Experimental Setup}

% overall setup, include picture of testing setup here with everything labeled
% for consistent naming, I'm calling the moving camera assembly the "Mobile MoCap," the flat plate at the end of the rail the "marker plate," 

For the automated evaluation of our Mobile MoCap system, we designed a robotic testing platform that consists of a servo-controlled camera assembly and a linear rail (LoPro), as shown in Fig.~\ref{fig:experimental_setup}. The linear rail, belt-driven with a high-precision stepper motor, allows for a means of reliably moving the Mobile MoCap assembly at a fixed rate with high repeatability. The rail's architecture is described further in~\cite{hanson2022occluded}. To track the position of the moving Mobile MoCap assembly relative to a static world frame, we used an OptiTrack V120:Trio (Natural Point), which recorded data at 120 Hz during all trials and serves as the ground truth for our experiments. Four 19~mm retroreflective marker spheres (Natural Point) were arranged on the opposite side of the acrylic plate to which the cameras are affixed for ground truth tracking with the OptiTrack. The OptiTrack Trio provides sub-millimeter precision, which is sufficient for evaluation. Mobile MoCap is intended as a compromise between more expensive static commercial rigs, such as the OptiTrack Trio, and inexpensive, less accurate fiducial marker approaches, like AprilTag. In order to assess the robustness of our system, we conducted experiments where the Mobile MoCap camera assembly was either static or moving (dynamic).

Using spherical (3D) retroreflectors instead of flat (2D) fiducial markers would provide a non-negligible advantage to the Mobile MoCap system in instances where the assembly's onboard cameras point at markers from non-orthogonal viewing angles. Therefore, we opted to test our system using only the flat, paper-based variant of retroreflectors to allow for a fair comparison against AprilTags. 

% To build a fair testbed for evaluation, we first custom-designed and laser-cut a flat backing plate to adhere both types of 2D markers. The backing plate was mounted to the distal end of the linear rail in a vertical orientation perpendicular to the rail's axis of motion. We produced three 30~mm-wide, circular markers from a sheet of retroreflective paper (Qualisys) using a die-cutting machine (Cricut) and arranged them asymmetrically on the backing plate (see Fig.~\ref{fig:title_teaser}a). Overall, the trio of discs covers a cumulative surface area of approximately 2120 $\text{mm}^2$. We also downloaded a set of AprilTags from the publicly-available tag36h11 family, randomly selected 1 square-shaped tag to test with (tag ID\#20), proportionally scaled it to be 52~mm wide, and printed it using a traditional office printer. In total, the AprilTag covered a surface area of 2704~$\text{mm}^2$. By forcing both styles of flat markers to live side-by-side on the same plane, we ensure comparable metrics between the two. In fact, given that the AprilTag is roughly 25\%  larger in surface area compared to the combined surface areas of the flat retroreflectors, we provide the AprilTag with a theoretical advantage in being seen by more of the RGB camera.

To build a fair testbed for evaluation, we laser-cut a flat backing plate out of acrylic. We then printed out an AprilTag Marker with a side length of 52~mm, and centered it on the acrylic plate. We cut three 30~mm circular markers from a sheet of retroreflective paper (Qualisys) using a die-cutting machine (Cricut) and arranged them asymmetrically on the plate, shown in Fig.~\ref{fig:title_teaser}a. These 30~mm markers are larger than those typically used with commercial systems. This is to ensure that the retroreflectors reflect light back into the stereo camera at larger distances where the Mobile MoCap ArduCam's integrated IR LEDs are weaker than those included with commercial IR-tracking cameras. In future work, Mobile MoCap will boast brighter IR LEDs in larger quantities, housed in a dedicated light source. We then mounted this plate to the distal end of the linear rail so that it was perpendicular to the table, and could be seen from the Mobile MoCap testing rig, shown in Fig.~\ref{fig:experimental_setup}.

\subsection{Static Trials}

As a base comparison point of pose estimation, we first tested the Mobile MoCap system at various distances and angles from the marker plate. The distances chosen were 0.9, 1.34, 1.78, and 2.23~m, and the angles chosen were 20, 10, and 0 degrees deflection from normal (yaw). The yaw rotation axis was specifically chosen for testing, as this most aligns with the motion in which wheeled robots operate.  Each angle was recorded for 2 seconds at 30~FPS on all three cameras (two cameras in IR for Mobile MoCap and one camera for AprilTag in RGB). This was repeated 30 times for each angle and distance, for a total of 120 trials. 

\subsection{Dynamic Angular Velocity Trials}

Moreover, we evaluate the system's robustness to rotational motion blur by rotating the Mobile MoCap assembly through a 40 degree range of motion with varying angular velocities. First, the Mobile MoCap assembly is fixed at a distance 1~m away from the marker plate. Then, the assembly is rotated through constant angular velocities of 0.05, 0.1, 0.2, and 0.4~rad/s. During these trials, the three cameras on the mobile assembly are recording at 30~Hz. Each angular velocity value is run 30 times, for a total of 120 trials. 

\subsection{Dynamic Linear Velocity Trials}

Akin to the previous set of angular velocity trials, we also test the tracking fidelity of our system while in translational motion. We moved the camera assembly back and forth along the rail at constant velocities of 10, 20, 25, and 30~cm/s, starting at 0.9~m and ending at 2.2~m. The overall process was repeated five times.  

\section{Results}
\label{sec:results}

To quantify estimation error in object tracking, we consider metrics related to both position and orientation. Position error is calculated as the root mean squared error (RMSE) between the OptiTrack's observed 3D coordinate $(x_1,y_1,z_1)_k$ (``ground truth'') and its corresponding estimate $(x_2,y_2,z_2)_k$ for sample $k$. Given that the OptiTrack operates at a significantly faster frequency than the AprilTag and Mobile MoCap tracking methods, every sequence of estimates had to be synchronized to a common rate. This synchronization procedure yields $n$ total samples and an RMSE metric on 3D position expressed as:
\begin{equation*}
    p_{\mathrm{rmse}} = \sqrt{\sum_{k=1}^{n}\frac{(x_{1,k}-x_{2,k})^2+(y_{1,k}-y_{2,k})^2 + (z_{1,k}-z_{2,k})^2}{n}}.
\end{equation*}

To evaluate 3D rotation error, we consider an inner dot product of the ground truth and estimated unit quaternions. This error function can be viewed as a distance between orientations~\cite{huynh2009metrics}. Let OptiTrack's ground truth frame orientation be $\mathbf{q}_{\mathrm{1},k}$, and the corresponding estimated quaternion be $\mathbf{q}_{\mathrm{2},k}$, then the error in rotation for sample $k$ is:
\begin{equation*}
    q_{\mathrm{err},k} = \frac{\arccos{(|\mathbf{q}_{\mathrm{1,k}} \cdot \mathbf{q}_{\mathrm{2,k}}|)}}{n},
\end{equation*}
where $q_{\mathrm{err},k}\in[0,\pi]$.

For each row of measurements reported in Tables~\ref{tab:static_april}-\ref{tab:dynamic_ang_vel}, the above metrics are averaged over multiple trials conducted for the same parameter set. The values for $x$, $y$, and $z$ are the mean positional error in each of the three axes plus-minus the standard deviation of that axis' relative error.

\subsection{Static}

\begin{table}[t]
\caption{AprilTags -- Static Trials Performance}
\label{tab:static_april}
\centering
\footnotesize
\setlength\tabcolsep{2.5 pt} % Essential for spacing
\begin{tabular}{cc|c @{} cccc @{} c}
\toprule
$d\,(\mathrm{m})$ & $a\,(\mathrm{rad})$ & $p_{\mathrm{rmse}}\,(\mathrm{cm})$ & $x\,(\mathrm{cm})$ & $y\,(\mathrm{cm})$ & $z\,(\mathrm{cm})$ & $q_{\mathrm{err}}\,(\mathrm{rad})$ \\
\midrule
\multirow{3}{*}{0.9} & 0.0 & 5.11 & -4.18 $\! \pm \!$ 0.1 & 7.46 $\! \pm \!$ 0.1 & -2.18 $\! \pm \!$ 0.6 & 0.19 \\
& 0.17 & 5.52 & -4.23 $\! \pm \!$ 0.3 & 7.67 $\! \pm \!$ 0.7 & 0.54 $\! \pm \!$ 3.7 & 0.18 \\ 

&-0.34 & 7.57 & -3.54 $\! \pm \!$ 2.1 & 9.02 $\! \pm \!$ 0.5 & 8.31 $\! \pm \!$ 2.0 & 0.11 \\ \hline

\multirow{3}{*}{1.3} & 0.0 & 8.82 & -1.05 $\! \pm \!$ 0.2 & 6.50 $\! \pm \!$ 0.1 & -13.67 $\! \pm \!$ 1.7 & 0.39 \\

 & 0.17 & 6.89 & -1.67 $\! \pm \!$ 1.4 & 6.57 $\! \pm \!$ 0.8 & -4.71 $\! \pm \!$ 8.5 & 0.36 \\ 

&-0.34 & 8.38 & -0.93 $\! \pm \!$ 1.8 & 8.19 $\! \pm \!$ 0.8 & 9.25 $\! \pm \!$ 7.3 & 0.15 \\ \hline

\multirow{3}{*}{1.8} & 0.0 & 16.61 & 2.80  $\! \pm \!$ 0.9 & 6.16 $\! \pm \!$ 0.4 & -26.62 $\! \pm \!$ 8.5 & 0.43 \\

 & 0.17 & 22.37 & 2.94 $\! \pm \!$ 3.4 & 6.91 $\! \pm \!$ 1.0 & -34.23 $\! \pm \!$ 16.2 & 0.42 \\ 

&-0.34 & 7.99 & -3.19 $\! \pm \!$ 2.9 & 8.54 $\! \pm \!$ 1.3 & -3.84 $\! \pm \!$ 9.2 & 0.31 \\ \hline

\multirow{3}{*}{2.2} & 0.0 & 53.43 & 9.29 $\! \pm \!$ 1.7 & 7.96 $\! \pm \!$ 0.6 & -90.40 $\! \pm \!$ 15.5 & 0.65 \\

 & 0.17 & 52.75 & 8.58 $\! \pm \!$ 4.1 & 8.04 $\! \pm \!$ 0.8 & -87.76 $\! \pm \!$ 22.2 & 0.64 \\
    
    & -0.34 & 16.48 & 1.98 $\! \pm \!$ 4.2 & 7.76 $\! \pm \!$ 1.2 & -0.07 $\! \pm \!$ 27.1 & 0.27  \\
    \bottomrule
    \end{tabular}
\end{table}

\begin{table}[t]
\caption{Mobile MoCap -- Static Trials Performance}
\label{tab:static_mobile_mocap}
\centering
\footnotesize
\setlength\tabcolsep{1.9 pt} % Essential for spacing
\begin{tabular}{cc|c@{}ccccc@{}c}
\toprule
$d\,(\mathrm{m})$ & $a\,(\mathrm{rad})$ & $p_{\mathrm{rmse}}\,(\mathrm{cm})$ & $x\,(\mathrm{cm})$ & $y\,(\mathrm{cm})$ & $z\,(\mathrm{cm})$ & $q_{\mathrm{err}}\,(\mathrm{rad})$ \\
\midrule
\multirow{3}{*}{0.9} & 0.0 & 1.91 & $-0.79 \pm 0.0$ & $6.48 \pm 0.0$ & $0.05 \pm 0.0$ & 0.15 \\
& 0.17 & 2.70 & $-0.48 \pm 0.7$ & $6.76 \pm 0.7$ & $1.83 \pm 4.1$ & 0.15 \\
& -0.34 & 6.39 & $0.85 \pm 2.2$ & $8.10 \pm 0.5$ & $10.22 \pm 2.3$ & 0.12 \\ \hline
\multirow{3}{*}{1.3} & 0.0 & 0.96 & $0.89 \pm 0.1$ & $4.65 \pm 0.1$ & $-2.67 \pm 0.1$ & 0.17 \\
& 0.17 & 1.97 & $1.05 \pm 1.5$ & $5.09 \pm 1.0$ & $-0.22 \pm 5.8$ & 0.17 \\
& -0.34 & 5.89 & $2.01 \pm 2.3$ & $6.89 \pm 0.9$ & $8.76 \pm 7.6$ & 0.18 \\ \hline
\multirow{3}{*}{1.8} & 0.0 & 0.34 & $3.05 \pm 0.1$ & $3.58 \pm 0.0$ & $-5.60 \pm 0.07$ & 0.17 \\
& 0.17 & 1.25 & $3.40 \pm 1.5$ & $4.01 \pm 1.1$ & $-3.67 \pm 10.8$ & 0.16 \\
& -0.34 & 6.51 & $3.39 \pm 1.8$ & $6.16 \pm 1.2$ & $9.98 \pm 7.7$ & 0.15 \\ \hline
\multirow{3}{*}{2.2} & 0.0 & 1.11 & $2.30 \pm 0.1$ & $2.56 \pm 0.1$ & $-8.20 \pm 0.4$ & 0.29 \\
& 0.17 & 0.60 & $2.50 \pm 2.6$ & $3.56 \pm 2.2$ & $-4.27 \pm 10.9$ & 0.42 \\
& -0.34 & 10.08 & $-11.05 \pm 8.3$ & $9.06 \pm 3.5$ & $-28.25 \pm 17.6$ & 0.93 \\
    \bottomrule
    \end{tabular}
\end{table}

The results in Tables~\ref{tab:static_april} and~\ref{tab:static_mobile_mocap} are promising for our proposed stereo retroreflector system. Every $p_{\mathrm{rmse}}$ value for Mobile MoCap is less than its corresponding value for AprilTags, with sub-two centimeter error at 0.9~m from the marker plate at zero degrees. One potential outlier is a $p_{\mathrm{rmse}}$ value of 0.6~cm at 2.2~m and 0.17~radians in Table~\ref{tab:static_mobile_mocap}, which is lower than the preceding value. Nonetheless, for every other set of three values at each distance with increasing angular deflection, the $p_{\mathrm{rmse}}$ value shows a positively increasing trend. At -.34~rad at 1.8~m/2.2~m, AprilTag outperforms Mobile MoCap in estimating the $z$-axis, although with larger variance. The $p_{\mathrm{rmse}}$ for Mobile MoCap is lower indicating better overall performance for Mobile MoCap. This anomaly in Mobile MoCap is likely due to the current lack of handling erroneously triangulated markers when determining pose under noisy scene conditions, such as when the camera is tilted further from the normal to the target. Future work will filter out erroneously triangulated markers to increase Mobile MoCap's effectiveness in more noisy scenarios.

\subsection{Dynamic Angular Velocity}

\begin{table}[t]
\caption{Dynamic Angular Velocity Trials Performance}
\label{tab:dynamic_ang_vel}
\centering
\footnotesize
\setlength\tabcolsep{4 pt} % Essential for spacing
\begin{tabular}{c|cccc}
\toprule
 \multirow{2}{*}{$a_{vel}\,(\mathrm{rad/s})$} & \multicolumn{2}{c}{AprilTag} & \multicolumn{2}{c}{Mobile MoCap} \\
 & $p_{\mathrm{rmse}}\,(\mathrm{cm})$ & $q_{\mathrm{err}}\,(\mathrm{rad})$  & $p_{\mathrm{rmse}}\,(\mathrm{cm})$ & $q_{\mathrm{err}}\,(\mathrm{rad})$  \\
\midrule
0.05 & 8.86 & $0.165 \pm 0.09$ & 3.88 & $0.156 \pm 0.05$ \\
0.1 & 9.07 & $0.130 \pm 0.07$ & 5.02 & $0.143 \pm 0.05$ \\
0.2 & 9.16 & $0.146 \pm 0.08$ & 5.53 & $0.142 \pm 0.05$ \\
0.4 & 8.90 & $0.162 \pm 0.10$ & 5.04 & $0.155 \pm 0.09$ \\
    \bottomrule
    \end{tabular}
\end{table}

The results in Table~\ref{tab:dynamic_ang_vel} show a significant trend in favor of Mobile MoCap. In all of the angular rotation settings, the Mobile MoCap system demonstrates smaller positional error than AprilTags. Although the improvement is marginal, the results were collected at 1 m distance, where the visual recognition of the AprilTag is still reliable. At further distances, up to the maximum range of the system, the Mobile MoCap performance would likely be superior. The orientation error is also notably lower for Mobile MoCap, although the performance increase at this distance is still within a standard deviation of the AprilTag. Across both localization systems, increasing $a_{vel}$ is directly correlated with an increase in pose error.

\subsection{Dynamic Linear Velocity}

\begin{table}[t]
\caption{Dynamic Linear Velocity Trials Performance}
\label{tab:dynamic_lin_vel}
\centering
\footnotesize
\setlength\tabcolsep{4 pt} % Essential for spacing
\begin{tabular}{c|cccc}
\toprule
 \multirow{2}{*}{$l_{vel}\,(\mathrm{cm/s})$} & \multicolumn{2}{c}{AprilTag} & \multicolumn{2}{c}{Mobile MoCap} \\
 & $p_{\mathrm{rmse}}\,(\mathrm{cm})$ & $q_{\mathrm{err}}\,(\mathrm{rad})$  & $p_{\mathrm{rmse}}\,(\mathrm{cm})$ & $q_{\mathrm{err}}\,(\mathrm{rad})$  \\
\midrule
10 & 9.09 & $0.170 \pm 0.10$ & 2.62 & $0.140 \pm 0.03$ \\
20 & 31.01 & $0.332 \pm 0.15$ & 1.60 & $0.178 \pm 0.07$ \\
25 & 14.93  & $0.190 \pm 0.12$ & 2.35 & $0.147 \pm 0.06$ \\
30 & 23.8 & $0.256 \pm 0.16$ & 2.10 & $0.145 \pm 0.07$ \\
    \bottomrule
    \end{tabular}
\end{table}

The values in Table~\ref{tab:dynamic_lin_vel} show a number of distinct trends. As the constant linear velocity values increase, the $p_{\mathrm{rmse}}$ values for AprilTags gradually increase, while the $p_{\mathrm{rmse}}$ values for Mobile MoCap remain more or less the same. This fruitful result suggests that the Mobile MoCap system is robust to motion blur. Additionally, the normalized $q_{\mathrm{rmse}}$ values associated with orientation error are on average larger for AprilTags than for Mobile MoCap. In general, these results demonstrate the full potential of the Mobile MoCap system as a mobile solution, whereby minimizing motion blur poses less risk of errors in localization. Lastly, it is worth noting that the standard deviations in position and orientation errors are significantly less than those of AprilTags.

% \begin{figure}[t]
%     \centering
%     \includegraphics[width=\columnwidth]{figures/mocap_results_comparison.png}
%     \caption{A series of boxplots showing the comparison between AprilTags and Mobile MoCap. The RMSE values for the position and rotation trials were measured in cm and rad respectively.}
%     \label{fig:mocap_results}
%     \vspace{-3.4mm}
% \end{figure}

\section{Discussion}

Fiducial markers, such as AprilTags~\cite{Olson2011April,Wang2016April2}, are limited by the minimum resolvable distance between points in an image. As AprilTags are based on a lexographic coding system, the detection pipeline aims to detect line segments and quads within a target area. At long distance, especially when the camera has low resolution or poor focus, the number of pixels occupied by a quad is small, thus decreasing the likelihood of detection. The Mobile MoCap system overcomes this challenge by relying on simpler geometric features that are recognized individually, rather than in the context of a larger grid encoding. While Mobile MoCap showed similar positional tracking performance when evaluated against AprilTags, there was a noticeable difference in dynamic angular tracking performance. This is likely due to the vast reduction in motion blur artifacts due to the reduced exposure time, which is a unique feature of Mobile MoCap. These experiments were performed to evaluate the viability of future tests involving a moving camera system and a moving target, which will be covered in future work extending Mobile MoCap in the muli-vehicle domain.

The Mobile MoCap system is not without limitations. Unlike AprilTags, there is no inherent association between marker configuration and object identifier (ID). It is possible to associate a marker geometry to a specific ID, but this is not yet a built-in feature, unlike existing fiducial marker libraries. Similar to other more expensive systems, our system is also subject to noise from highly reflective surfaces, such as metallic screw heads or reflective trim on articles of clothing. These items can be falsely detected as features at close distances. Most indoor lighting is provided by LED or fluorescent lights spanning the visible light spectrum, resulting in low noise from indoor lights. However, the sun provides a broader spectrum, and incidental sunlight can be falsely detected as a marker. The current iteration of this system is best suited for indoor use, or outdoor use during the night. Nevertheless, our results demonstrate that our system is functional and accurate in dynamic environments where more expensive commerical motion tracking systems optimally operate.

\section{Conclusions}

In this work we presented Mobile MoCap, a novel motion capture system architecture that is both inexpensive and portable. We demonstrated the system's superior performance over a popular fiducial marker approach, AprilTags, in terms of object 6-DoF localization accuracy. In future iterations of this work, we plan to exchange the camera system for a system capable in the Short Wave IR range, allowing for robust performance both indoors and outdoors. Additionally, this system will be mounted to a variety of mobile robots to explore multi-vehicle tracking. Our system and methods empirically demonstrate motion capture systems can be made in cost-effective formats, while still providing excellent localization accuracy in robotics contexts.


%%%%%%%%%%%%%%%%%%%%%%%%%%%%%%%%%%%%%%%%%%%%%%%%%%%%%%%%%%%%%%%%%%%%%%%%%%%%%%%%

%%%%%%%%%%%%%%%%%%%%%%%%%%%%%%%%%%%%%%%%%%%%%%%%%%%%%%%%%%%%%%%%%%%%%%%%%%%%%%%%

%%%%%%%%%%%%%%%%%%%%%%%%%%%%%%%%%%%%%%%%%%%%%%%%%%%%%%%%%%%%%%%%%%%%%%%%%%%%%%%%

\section*{Acknowledgment}
The authors would like to thank Alina Spectorov for aid in debugging the Mobile MoCap software. This research has been supported in part by Defense Advanced Research Projects Agency (DARPA) under HR0011-22-2-0004. The views expressed are those of the authors and do not reflect the official policy or position of the Department of Defense or the U.S. Government. Approved for Public Release, Distribution Unlimited.

%%%%%%%%%%%%%%%%%%%%%%%%%%%%%%%%%%%%%%%%%%%%%%%%%%%%%%%%%%%%%%%%%%%%%%%%%%%%%%%%

\bibliographystyle{IEEEtran}
\bibliography{IEEEabrv,references}
\addtolength{\textheight}{-12cm}   % This command serves to balance the column lengths

\end{document}